%% file: iclr2025_re-align_workshop.tex
\documentclass{article} 
\usepackage{iclr2025_re-align_workshop,times}

\input{math_commands.tex}

\usepackage{hyperref}
\usepackage{url}
\usepackage{graphicx}
\usepackage{subcaption} 
\usepackage{makecell}
\usepackage{booktabs}
\usepackage{multirow}
\usepackage{hhline}

\title{Cognitive Neural Architecture Search \\ Reveals Hierarchical Entailment}

\author{Lukas Kuhn$^{*}$, Sari Saba-Sadiya$^{*,\dagger}$, Gemma Roig \\
Goethe University Frankfurt, Department of Computer Science,  Frankfurt, Germany \\
The Hessian Center for Artificial Intelligence (hessian.AI), Darmstadt, Germany \\
}

\iclrfinalcopy 
\begin{document}

\maketitle

\begin{abstract}
Recent research has suggested that the primate brain is more shallow than previously thought, challenging the traditionally assumed hierarchical structure of the ventral visual pathway. Here, we demonstrate that optimizing convolutional network architectures for brain-alignment via evolutionary neural architecture search results in models with clear representational hierarchies. Despite having random weights, the identified models achieve brain-alignment scores surpassing even those of pretrained visual classification models - as measured by both linear encoding and representational similarity analysis. Furthermore, architectures optimized for alignment with late ventral regions perform at the level, or better, than state-of-the art models when trained on image classification tasks. These findings suggest that hierarchical structure is a fundamental mechanism of primate visual processing. Finally, this work demonstrates the potential of neural architecture search as a framework for computational cognitive neuroscience research that could reduce the field's reliance on manually designed convolutional networks.
\end{abstract}

\section{Introduction}


\def\thefootnote{*}\footnotetext{These authors contributed equally to this work}\def\thefootnote{\arabic{footnote}}
\def\thefootnote{$\dagger$}\footnotetext{Corresponding Author: \texttt{sadiya@rz.uni-frankfurt.de}\\}\def\thefootnote{\arabic{footnote}}

Throughout the last decade, Convolutional Neural Networks (CNNs) emerged as powerful cognitive models capable of providing valuable insight into the neural mechanisms underlying primate visual processing \citep{Yamis2014,StYves2022,Manshan2025}. In their seminal work, \cite{Yamis2014} demonstrated that CNNs trained to perform image classification can be used to predict brain activity with greater accuracy than previously developed cognitive models. Moreover, their findings suggested a shared representational hierarchy between CNN layers and visual cortex regions, where intermediate and late CNN layers correspond to intermediate and late visual processing regions, respectively. However, recent research that directly explored the emergence of brain-like hierarchy in neural networks trained to directly predict brain activity found evidence against the necessity of entailment hierarchy \citep{StYves2022}. Based on these results, the authors posit the \textit{shallow brain hypothesis}, arguing that low-level representations may not be necessary preprocessing stages for higher-level representations. Our work expands on this \textit{shallow vs deep-brain} debate by employing Neural Architecture Search (NAS) to explore the emergence of early visual cortex-like representation in network architectures optimized to align with late ventral representation. 

Previous studies have demonstrated that it is possible to identify CNNs with state-of-the-art classification performance by directly optimizing model architectures using methods such as reinforcement learning or genetic algorithms. For instance, \citep{GeneticCNN,liu2018hierarchical} leveraged genetic algorithms to `evolve' architectures that outperform manually designed CNNs on MNIST and CIFAR-10 datasets. More recently, \cite{mundt2021neuralarchitecturesearchdeep} employed NAS to identify network architectures that, even without gradient descent training, compute representations that enable classification performance comparable to fully trained deep networks by simply training a linear probe to predict the image label. Building on this, we optimized CNN architectures to predict cognitive representations across different regions of the ventral stream. We formulate a simple hypothesis in favor of the \textit{deep-brain} model: Optimizing CNNs to predict late visual representations in the inferior temporal (IT) cortex would spontaneously optimize their alignment with representations found in the early (V2) and intermediate (V4) visual cortex regions in lower layers. Our results demonstrate that NAS can identify CNN architectures with better brain alignment than manually designed image classification models such as AlexNet, VGG16 and CORNet. Furthermore, we found that the optimal CNNs for predicting V2 and V4 representations were sub-networks of those optimized for predicting IT representations, indicating that these earlier representations might be necessary when trying to predict high level visual representations, therefore providing evidence in favor of the \textit{deep-brain} model of the visual cortex. Finally, we also observed that architectures optimized to predict brain representations achieved competitive results when trained for image classification.

\begin{figure}[t]
    \centering
    {\includegraphics[width=12cm]{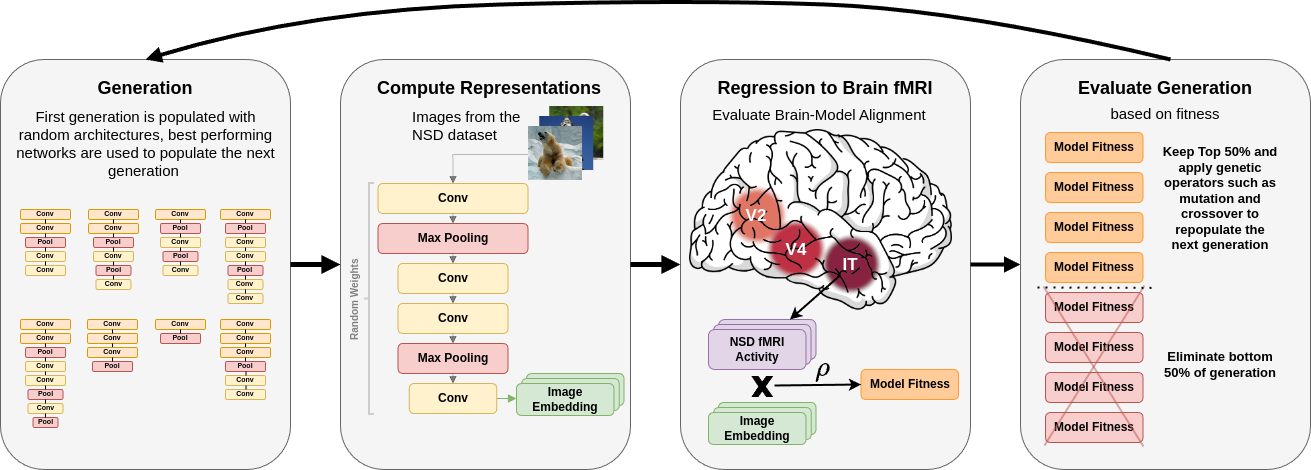}}
    \caption{The evolutionary neural architecture search framework: Starting with a generation of neural architectures, for each model embeddings of all images in the shared NSD dataset are extracted. A ridge regression is then trained to predict the recorded fMRI activity and the correlation coefficient between the predicted and ground truth fMRI is calculated. The models are evaluated based on the mean correlation for all subjects, and the bottom 50\% are eliminated. Finally genetic operations are used to repopulate the models for the next generation.}
    \label{fig:accuracy}
    \vspace{-1em}
\end{figure}

\section{Method}
 We follow the standard evolutionary NAS methodology of allowing the genetic algorithm to perform selection, mutation, and crossover of the best networks in each generation to find an optimal architecture for brain alignment \citep{liu2018hierarchical}. In the following section, we present the specifics of this evolutionary architecture search. That is, we discuss the search space, the evaluation strategy, and the genetic operators used to evolve each new generation of CNNs. 

\subsection{Search Space}
Following \cite{GeneticCNN}, we construct an initial generation of random individual networks, where each is a standard hierarchical stack of multiple convolution and max-pooling layers. We empirically found that adding linear layers to the search space did not improve brain-alignment and therefore excluded them to allow for faster convergence. We also limit the searchable hyper-parameter ranges of the convolution layers kernel size (3 to 11), stride (1 to 4) and number of filters (64 to 512). Furthermore, the max-pooling layers kernel size range was 2 to 3. Finally, to further restrict the search space we also enforce the CNNs to always have a monotonically increasing number of filters across the layers, as receptive fields are expected to expand to integrate more conceptual information in deeper layers.


\subsubsection{Fitness Evaluation}
Adapting the approach of \cite{mundt2021neuralarchitecturesearchdeep} we evaluate multiple randomly initialized versions of the same network, guaranteeing that the models are picked based on architecture, rather than a lucky weight initialization (the lottery ticket hypothesis). We use the Net2Brain toolbox to evaluate network performance \cite{net2brain}: First, we extract the last layer image encoding and train a ridge regression to predict fMRI responses to the same image. We then calculate the Pearson correlation coefficient of predicted and ground-truth neural responses on a held-out test set. The coefficient is averaged across ten consecutive random seeds for each subject to generate model fitness. Finally, to speed convergence, in the first two generations we artificially increase the population size by evaluating every layer of each network (not just the last) and use the best performing sub-network in following generations. Following common practice, the random network weights were initialized using a Kaiming uniform distribution with a bias of zero.

\subsection{Genetic Operators}
After all the models in a generation are evaluated, we remove the bottom 50\% of the CNN population based on fitness. We then create new offspring networks to repopulate the population. This is achieved using standard mutation and crossover genetic operators \citep{GeneticCNN}.

Our mutation strategy employs three operators on each selected parent network: addition, modification, and removal. The addition operation introduces new layers while maintaining architectural validity, with special attention to channel dimensionality progression and layer-type constraints. Modification alternate between layer-type transformations (with probability $P=0.3$) and parameter refinements ($P=0.7$) in which architecture elements such as the kernel size are adjusted. The removal operation preserves network integrity by selectively eliminating layers while maintaining essential architectural elements (minimum depth of one layer and output size larger than one).

Our crossover operator implements a single-point crossover strategy in which architectures exchange structural information at a randomly selected position. The operation creates offspring by preserving the parent's layers up to the crossover point and inheriting the remaining layers from the second parent, maintaining architectural validity through constraint checking.

\subsection{Dataset}
The regression was trained using data from the NSD Dataset \cite{NSDDataset}, a large-scale fMRI dataset of 8 subjects viewing thousands of natural scenes. Specifically, we used a set of $872$ images shared across all subjects. Following \citep{Manshan2025}, we used a subset of five subjects (subjects 1,2,4,5,7) with a high signal-to-noise ratio (SNR). Moreover, we focused on the brain activity recorded in V2, V4, and the Inferior Temporal cortex (IT) as stand-ins for representations in the early, intermediate, and late ventral stream respectively. Specifically, IT activations were constructed by concatenating activations from the FFA, FBA, EBA and PPA regions \cite{kanwisher2013functional}. 

\subsection{Baseline Models}
We compare the brain-alignment of our models against well known CNNs such as AlexNet and VGG16 which are often used as cognitive models (see \cite{Yamis2014,Manshan2025}). Moreover, we also use CORNetS which was specifically designed to process information in a more brain-like manner \cite{CORNet}.

\begin{figure}[t]
    \centering
    {\includegraphics[width=12cm]{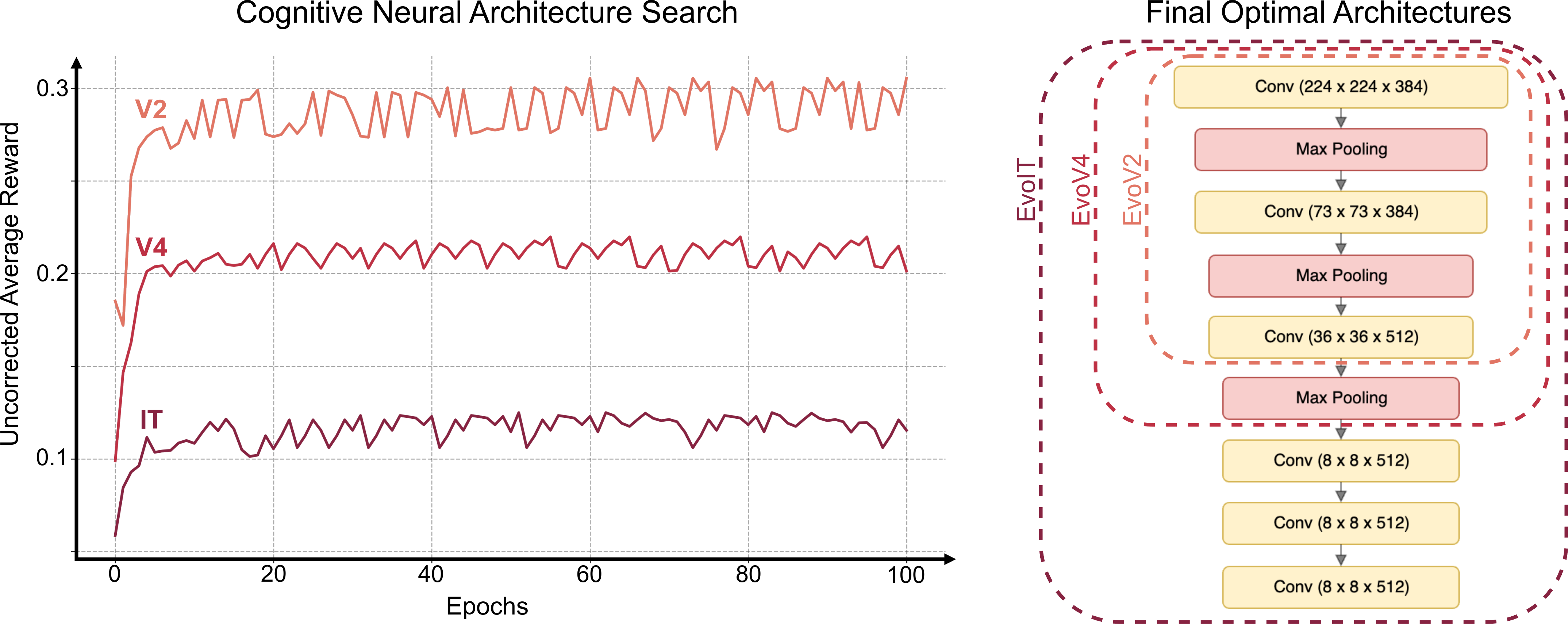}}
    \caption{Left: The average rewards for the architectures optimized to predict V2, V4, and IT brain activations in each generation. Right: The optimal \emph{EvoIT} architecture.}
    \label{fig:combined}
    \vspace{-1em}
\end{figure}

\section{Experiments and Results}
The brain alignment results reported in this section are in terms of percent of variance explained relative to the lower noise ceiling. For representational similarity analysis we followed the standard lower noise ceiling calculation \cite{RSA}. For the linear encoding based alignment score we used the regression method presented by \cite{LageCastellanos2019}. We report performance for model layer with the highest alignment score.

\subsection{Evolutionary Search for Cognitive Models}
We ran our evolutionary search three times for each brain region. Each run consisted of $100$ generations with a mutation rate of $0.25$ and a crossover rate of $0.5$. The optimal architectures discovered were identical in all runs optimizing for the alignment with the same brain region save for minor differences in stride sizes, indicating that the identified networks are robust to noise in the evolutionary search process. Furthermore, the three optimal architectures for V2, V4, and the IT - which we coin EvoV2, EvoV4, and EvoIT - had high alignment with their respective brain regions, outperforming most baseline models (Table \ref{tab:model-comparison}). The number of layers in the EvoV2, EvoV4, and EvoIT architectures was five, six, and nine respectively (Figure \ref{fig:combined} right).

\begin{table}[ht]
\centering
\renewcommand{\arraystretch}{1.2}
\begin{tabular}{|m{4em}|c|c|c|c|c|c|c|}
\hhline{~~------}
 \multicolumn{2}{c}{} & \multicolumn{2}{|c|}{V2} & \multicolumn{2}{c|}{V4} & \multicolumn{2}{c|}{IT} \\
 \hhline{~~------}
 \multicolumn{2}{c|}{} & Reg & RSA & Reg & RSA & Reg & RSA \\
\hline
\multirow{2}{=}{\centering AlexNet} & Random & 8 & 5.15 & 5.14 & 3.09 & 1.29 & 0.40 \\
& Trained & 4.6 & 14.51 & 2.59 & 7.54 & 1.12 & 1.16 \\
\hline
\multirow{2}{=}{\centering VGG16} & Random & 1.09 &  3.03 &  5.4 & 1.51  & 0.95  &  0.12 \\
& Trained & \textbf{13.76} & 16.12 & 4.82 & 6.91 & 0.75 & 0.85 \\
\hline
\multirow{2}{=}{\centering CORNet} & Random & 2.38 & 1.52 & 2.33 & 0.91 & 0.65 & 0.33 \\
& Trained & 2.276 & \textbf{17.82} & 1.4 & \textbf{9.99}  & 1.13 & 1.09 \\
\hline
\multirow{2}{=}{\centering EvoV2} & Random & 11.9 & 5.99 & 6.15 & 3.47 & 0.95 & 0.41 \\
& Trained & 3.76 & 6.33 & 1.81 & 3.52  & 0.69 & 0.56 \\
\hline
\multirow{2}{=}{\centering EvoV4} & Random & 11.98 & 3.94 & 6.5 & 2.47 & 1.2 & 0.41 \\
& Trained & 3.544 & 6.32 & 1.9 & 4.00 & 0.55 & 0.60 \\
\hline
\multirow{2}{=}{\centering EvoIT} & Random & 10.64 & 3.90 & \textbf{6.63} & 2.70 & \textbf{1.92} & 0.43 \\
& Trained & 8.82 & 16.33 & 4.06 & 8.27 & 1.04 & \textbf{1.18} \\
\hline
\end{tabular}
\caption{Brain-model similarity as measured by linear encoding (Reg) and representational similarity analysis (RSA) across brain regions and model architectures with trained and random weights. Similarity is given calculated by normalizing the correlation coefficient using the lower noise ceiling. All Trained models use weights optimized on the CIFAR-10 classification (Section \ref{sub:cifar}).}
\label{tab:model-comparison}
\vspace{-1em}
\end{table}

\subsection{Representational Hierarchy in Evolutionary Cognitive Models}
We formulated the following hypothesis in favor of the existence of hierarchical entailment across brain region representations: optimizing CNNs to predict activations in the IT representations will also spontaneously optimize the networks to learn V2 and V4 representations. To accomplish this we tested the brain-alignment of each brain region with each architecture (Table \ref{tab:model-comparison}). Indeed, we find that EvoIT contained sub-networks that are competitive predictors of V2 and V4 (in fact, a subnetwork of EvoIT was the best in class V4 model). Specifically, the EvoIT layer that was most correlated with V4 was the third pooling layer (Figure \ref{fig:combined} right). Moreover, manually inspecting the best performing architectures we observed that the EvoV4 and EvoIT architectures contained subnetworks virtually identical to EvoV2 and EvoV4 respectively. Overall, these results indicate that to compute representations similar to those found in the IT it is indeed beneficial to first compute representations similar to those found in V2 and V4.  

\subsection{Training Evolutionary Cognitive Models for Image Classification} \label{sub:cifar}

CNNs designed to perform image classification are state-of-the-art cognitive models \cite{Yamis2014}. Here we investigate if CNNs with architectures that were specifically optimized for brain-alignment can be trained to perform image classification. To achieve this we used the best architectures identified by the evolutionary search by adding a linear layer with Softmax on top of the CNN backbone and training the models with cross-entropy on the CIFAR-10 dataset for 7 epochs. To establish multiple baselines we also randomized the weights of the baseline CNNs and trained them using a similar procedure.

We observe that the classification performance increased for models optimized to align with later ventral regions, with EvoV2 having the lowest score, followed by EvoV4 and EvoIT. Overall, the performance of EvoIT was close to the performance of several baseline models (Table \ref{tab:cifar10}). 

\begin{table}[ht]
\centering
\begin{tabular}{ccccccc} \toprule
 \rule{0pt}{2ex} & EvoV2 & EvoV4 & EvoIT & VGG16 & AlexNet & CORNet \\
\midrule
 \makecell{Top-1 \\ Accuracy} & 64.1 & 67.7 & 77.2 & 78.5 & 79.5 & 80.85 \\
\bottomrule
\end{tabular}
\caption{Classification performance after training randomly initialized models on CIFAR-10}
\label{tab:cifar10}
\vspace{-1em}
\end{table}  



\section{Discussion and Future Work}

In this paper we used genetic algorithm based neural architecture search to optimize the architecture of the convolutional neural networks directly to be more brain-like. Through this framework we identified network architectures that had high similarity to various brain regions despite the lack of any gradient descent based training. Moreover, we found that the model optimized for similarity with late ventral stream areas contained subnetworks that were virtually identical to those identified as the optimal models for similarity with early and intermediate visual cortex representations. This finding directly contributes to the recent discussion regarding the hierarchy  - or lack thereof - found across visual cortex representation \citep{Yamis2014,StYves2022}.

More broadly, the framework presented here is a potential new useful tool for computational cognitive neuroscience research. Previous research tackling questions regarding the architecture of the brain often followed a specific recipe: CNN architectures are modified in a controlled manner while the training data and function are held constant. Model-brain alignment is then measured to determine if the modification improves brain similarity, which would be taken as evidence that the modification constitutes an abstraction of a mechanism found in the brain (for example see \cite{Manshan2025}). In contrast, the cognitive NAS framework presented here optimizes the model architecture directly, instead of relaying on handcrafted CNNs, which might introduce unwanted bias. 

The initial results presented here highlight the potential utility of Cognitive NAS for the research community. However, the results of our experiments also raise multiple questions. Specifically, while the evolved networks were powerful encoding models, their performance was subpar when measured through representational similarity analysis. This might be due to our choice of using ridge regression for fitness evaluation. Interestingly, training the networks using the CIFAR-10 classification task improved brain-alignment as measured by RSA while lowering the regression score. This might indicate that the representation space of the evolved models lacks some structural elements that can only be learned through gradient descent training. Future work should carefully investigate the impact of the brain-similarity measure used during the NAS on the trajectory of the architecture search and the final networks identified as optimal cognitive models.   

\section*{Acknowledgements}
This work was partly funded by the German Research Foundation - DFG Research Unit FOR 5368.
\newpage
\bibliography{iclr2025_conference}
\bibliographystyle{iclr2025_conference}


\end{document}

%% file: math_commands.tex
\usepackage{amsmath,amsfonts,bm}

\def\eqref#1{equation~\ref{#1}}

\def\1{\bm{1}}

\DeclareMathAlphabet{\mathsfit}{\encodingdefault}{\sfdefault}{m}{sl}
\SetMathAlphabet{\mathsfit}{bold}{\encodingdefault}{\sfdefault}{bx}{n}

